\documentclass[a4paper]{svproc}
\usepackage{url}
\usepackage{graphicx}

\usepackage{amsmath,amssymb}
\usepackage{graphicx}
\usepackage{soul}
\usepackage{adjustbox}
\usepackage{booktabs}
\usepackage{multirow}
\usepackage{verbatim}
\usepackage{xcolor}
\usepackage{xspace}

\newcommand{\benchmark}{HRIBench\xspace}

\begin{document}
\mainmatter              
\title{\benchmark: Benchmarking Vision-Language Models for Real-Time Human Perception \\ in Human-Robot Interaction}
\titlerunning{\benchmark}  
%
\author{Zhonghao Shi \and Enyu Zhao \and Nathaniel Dennler \and Jingzhen Wang \and \\Xinyang Xu \and Kaleen Shrestha \and Mengxue Fu \and Daniel Seita \and Maja Matari\'c}
\authorrunning{Zhonghao Shi et al.} 
%
\tocauthor{Ivar Ekeland, Roger Temam, Jeffrey Dean, David Grove,
Craig Chambers, Kim B. Bruce, and Elisa Bertino}
\institute{
Thomas Lord Department of Computer Science, Viterbi School of Engineering\\
University of Southern California, Los Angeles CA 90089, USA
}

\maketitle              

\vspace{-3pt}
\begin{abstract}

Real-time human perception is crucial for effective human-robot interaction (HRI). Large vision-language models (VLMs) offer promising generalizable perceptual capabilities but often suffer from high latency, which negatively impacts user experience and limits VLM applicability in real-world scenarios. To systematically study VLM capabilities in human perception for HRI and performance-latency trade-offs, we introduce \textbf{\benchmark}, a visual question-answering (VQA) benchmark designed to evaluate VLMs across a diverse set of human perceptual tasks critical for HRI. \benchmark{} covers five key domains: (1) non-verbal cue understanding, (2) verbal instruction understanding, (3) human-robot-object relationship understanding, (4) social navigation, and (5) person identification. To construct \benchmark, we collected data from real-world HRI environments to curate questions for non-verbal cue understanding, and leveraged publicly available datasets for the remaining four domains. We curated 200 VQA questions for each domain, resulting in a total of 1000 questions for \benchmark. We then conducted a comprehensive evaluation of both state-of-the-art closed-source and open-source VLMs (N=11) on \benchmark. Our results show that, despite their generalizability, current VLMs still struggle with core perceptual capabilities essential for HRI. Moreover, none of the models within our experiments demonstrated a satisfactory performance-latency trade-off suitable for real-time deployment, underscoring the need for future research on developing smaller, low-latency VLMs with improved human perception capabilities. \benchmark and our results can be found in this Github repository: https://github.com/interaction-lab/HRIBench.

\end{abstract}

\section{Introduction}

With advances in robotics, research is increasingly shifting from constrained in-lab settings to in-the-wild environments, such as homes, where robots must effectively interact, assist, and collaborate with humans~\cite{gupta2018robotlearninghomesimproving,intelligence2025pi05,liu2024okrobot}. This shift requires robots to perceive human behaviors while reasoning about underlying human intentions, beliefs, and goals. Advances in deep-learning based computer vision (CV) have enabled the development of specialized models for visual HRI tasks~\cite{robinson2023robotic}. For instance, pointing gesture recognition models identify the location a human is pointing to in order to guide robot movements~\cite{nickel2007visual}, user engagement models assess a user's engagement level to determine social actions~\cite{ben2017ue}, facial recognition models identify individuals to adapt its actions to each individual user~\cite{cruz2008real}, and action recognition models interpret fine-grained human actions during interaction to determine appropriate assistive responses~\cite{zhdanova2020human}. Despite their strong performance, these specialized models are closed-set recognizers: they lack generalization abilities across different HRI tasks and cannot recognize novel or out-of-distribution classes at test time. Recent advancements in vision-language models (VLMs) have demonstrated impressive zero-shot capabilities, offering the potential for more general human perception for HRI. However, VLMs have not yet been systematically evaluated on human perception tasks in the context of HRI.

\begin{figure}[t!]
  \centering
  \includegraphics[width=\columnwidth]{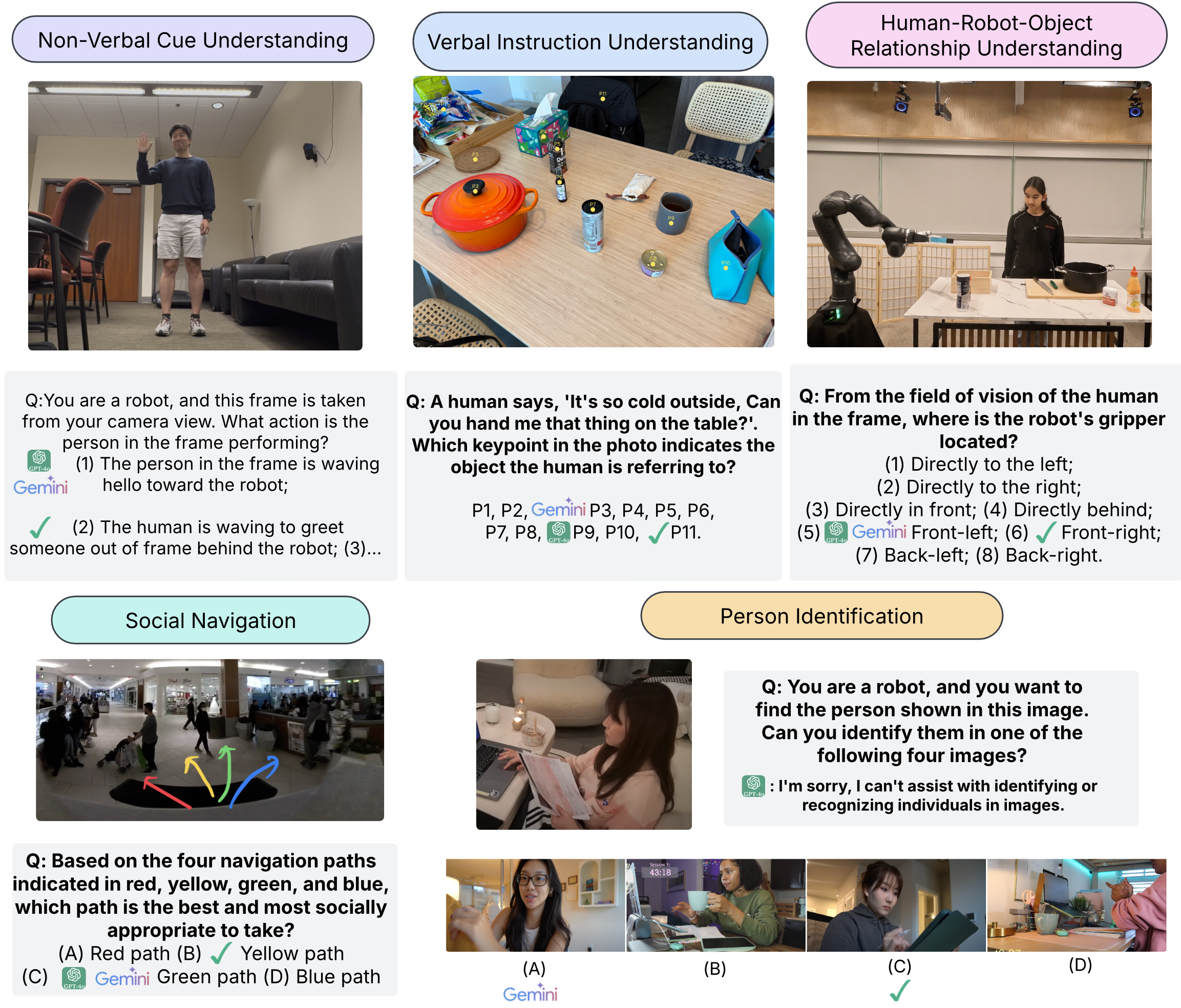}
  \caption{An example question for each of the five domains in \benchmark, along with evaluations from the GPT-4o and Gemini-2.5-pro VLMs. The green GPT and Gemini logos indicate their respective model selections, and the green check mark denotes the ground truth. Despite the internet-scale knowledge inherent in GPT-4o and Gemini-2.5-pro, both models lacked fundamental visual capabilities essential for real-world HRI, as demonstrated by their mistakes on these questions.}
  \label{fig:example}
\end{figure}

Prior work in robotics and machine learning proposed various benchmarks~\cite{li2025benchmark} to study VLMs' visual understanding for spatial reasoning~\cite{chow2025physbench} and robot manipulation~\cite{chen2025robo2vlm,guruprasad2024benchmarkingvisionlanguage,zhao2025ManipBench}. However, to our knowledge, no prior work has benchmarked VLMs for visual tasks in HRI or studied the performance-latency tradeoff for real-time inference in the context of HRI. This is an important topic because robot failures and delays in human perception can lead to user distrust and to misalignment between humans and robots~\cite{hancock2011meta}, ultimately undermining HRI success. Our work aims to address two key research questions toward advancing experimental HRI: \textit{(RQ1) To what extent can vision-language models (VLMs) enable accurate and generalizable human perception across different HRI tasks?} and \textit{(RQ2) What are performance-latency tradeoffs for state-of-the-art VLMs?}


To investigate these questions, we introduce \textbf{\benchmark}, a visual question-answering (VQA) benchmark that consists of five domains, as illustrated in Fig.~\ref{fig:example}. We conducted data collections in real-world HRI environments and also leveraged existing open-source datasets~\cite{wan2022handmethat,Kedia2023ManiCastCM,nguyen2023toward} to curate 200 novel VQA questions for each domain for \benchmark with a total 1000 questions. Our results show that both closed-source and open-source VLMs in our experiments still struggle to perceive human behaviors effectively with low latency. In particular, these models lack key capabilities such as understanding fine-grained multimodal cues (e.g., eye gaze), resolving ambiguous language-visual instructions, and performing real-world spatial and physical reasoning. 







\section{Methods}

Based on a recent survey~\cite{robinson2023robotic}, we categorized perceptual tasks in human-robot interaction (HRI) into five key domains, and created 200 novel VQA questions for each of the domains. Fig.~\ref{fig:example} shows an example question in each domain. In this section, we discuss each of the domains and how \benchmark questions are curated for each. 

\textbf{Non-verbal cue understanding.} Non-verbal cues such as hand gestures, body pose, pointing, and eye gaze are fundamental to human-robot communication~\cite{robinson2023robotic}. Although a substantial body of work in HRI has studied non-verbal cues extensively~\cite{urakami2023nonverbal}, there is no comprehensive open-source dataset that includes a diverse set of such cues to benchmark the perceptual capabilities of VLMs. To address this gap, we recorded a diverse range of non-verbal cues in HRI contexts. Two authors of this paper acted as actors and followed a standardized script to perform non-verbal cues under two conditions: (1) interacting with a robot and (2) interacting with a human. The goal of this domain is to evaluate not only whether models can recognize non-verbal cues, but also whether they can distinguish whether the cues are directed at the robot or at another person, a critical challenge in multi-party HRI. In the robot interaction condition, actors interacted with two robots (Kuri~\cite{kuri_robot} and Quori~\cite{specian2021quori}) of different heights; the varied camera angles introduced perceptual challenges. In the human interaction condition, actors engaged with another research team member standing beside the robot but outside its camera frame. This reflects a realistic HRI scenario in which robots often have limited visual fields and must reason about out-of-frame humans and objects. Based on the most commonly studied non-verbal cues identified by Robinson et al.~\cite{robinson2023robotic}, our script instructed the actors to perform 24 hand gestures (e.g., counting numbers, the OK sign, and thumbs-up/down), 17 body pose gestures (e.g., shrugging, bowing, and handing over objects), point in five directions, and shift their eye gaze at five different angles.

\textbf{Verbal instruction understanding.} Verbal instructions from humans can be ambiguous or may include incorrect information, requiring robots to reason using visual input and contextual information. Prior work in this area includes the HandMeThat benchmark~\cite{wan2022handmethat} that evaluates ambiguous instruction understanding in human-robot communication. HandMeThat provides a textual interface but does not use visual input. With \benchmark, we extend the ambiguous verbal instruction understanding task to include visual information. Using both visual input and textual context, VLMs were prompted to infer the human's intent to help achieve the overall task goal.

\textbf{Human-robot-object understanding.} In HRI, robots need to understand their physical position relative to humans and objects in the environment and they also need to take the human’s perspective to reason about their relative positions and surrounding objects. Establishing a shared coordinate system between the robot and the human is essential for effective communication and goal alignment. In this domain, we leveraged recordings from CoMaD~\cite{Kedia2023ManiCastCM} to create VQA questions that assess human-robot-object relationship understanding.

\textbf{Social navigation.} Robot navigation in crowded public spaces with humans is one of the most challenging fundamental tasks in HRI~\cite{mavrogiannis2023core}. Recent work has explored VLM capabilities as evaluators for path planning in navigation~\cite{aghzal2024evaluating}, but not yet in the context of robot social navigation. In this work, we utilized the most recent large-scale real-world social navigation dataset, MuSoHu~\cite{nguyen2023toward}, overlaid navigation paths of different colors onto the camera frame, and assessed the VLMs' ability to act as an evaluator in selecting the most socially appropriate path in a VQA format.



\textbf{Person identification.} Robots must be able to recognize human identities in order to personalize their interactions with individuals in settings such as the home, workplace, and schools. To evaluate the person identification capabilities of VLMs in these contexts, we used video frames extracted from publicly available YouTube ``Study/Work with Me" videos, where creators record themselves studying or working for extended periods. We sampled 20 video frames from 7 YouTubers. All selected YouTubers had publicly disclosed their age, allowing us to confirm that they were adults at the time the videos were recorded. From the sampled frames, we selected one as the query image (see Fig.~\ref{fig:example}), and one additional frame from the same YouTuber as the ground truth answer. We then randomly selected three frames from other YouTubers as distractor options for a multiple-choice question. These four frames were concatenated into a single image displaying the A/B/C/D answer options. Finally, both the query image and the concatenated options image were provided as input to the VLMs, along with the question prompt for model evaluation.



\section{Results and Discussion}
\label{sec:results}

We evaluated state-of-the-art (SOTA) closed-source ($N=6$)~\cite{Gemini,openai2024gpt4ocard} and open-source ($N=5$) VLM families across different model sizes~\cite{chen2024internvl,grattafiori2024llama} on the \benchmark dataset and compared results with human and random baselines, as shown in Table~\ref{tab:results}. For human evaluation, each domain was independently evaluated by a different human expert.

\begin{table}[t]
\scriptsize
\caption{Comparison of the human baseline, closed-source VLMs, open-source VLMs, and the random baseline on \benchmark. We report VLM accuracy across all five \benchmark domains, with average accuracy and latency (seconds per question) calculated across domains, excluding person identification. For person identification, N/A is reported for closed-source models from OpenAI (o3, GPT-4o, and GPT-4o-mini) due to their refusal to answer, and for the open-source Llama3.2 model due to its lack of support for multi-image input.}
\label{tab:results}
\begin{center}
\resizebox{0.99\columnwidth}{!}{%
\begin{tabular}{lc|ccccc|cc}
\toprule
\textbf{Model} & \textbf{\shortstack{\# of\\params.}}& \textbf{\shortstack{Relationship}} &  \textbf{\shortstack{Identification}} &  \textbf{\shortstack{Verbal}}   & \textbf{\shortstack{Non-Verbal}}   & \textbf{\shortstack{Navigation}} & \textbf{\shortstack{Avg.\\Acc.}} & \textbf{\shortstack{Avg.\\Latency}} \\
\midrule

Human               & N/A  & 0.90  & 0.98 & 0.96 & 0.98  & 0.96  & 0.93 & N/A\\
\midrule
\textbf{Closed-Source }             &  &  &  &  &  & &\\
o3                  &  N/A  &	\textbf{0.42}    &	N/A        &	0.74        &  \textbf{0.48}      &  \textbf{0.36}    &       \textbf{0.50}     &  41.64  \\
GPT-4o              &  N/A  &	 0.35      &	N/A        &	0.68        &   0.44     &  0.29     &  0.44          &  13.83   \\
GPT-4o-mini         &  N/A  &	0.22       &	N/A        &	0.39        &   0.36     &  0.26     &  0.30          &  12.21  \\
Gemini-2.5-pro      &  N/A  &	0.38       &	\textbf{0.82}        &	 \textbf{0.83}       &   0.45     &  0.32     &  0.49          &  19.33  \\
Gemini-1.5-pro      &  N/A  &	0.31       &	0.69        &	 0.64       &   0.38     &  0.30     &  0.40          &  2.66  \\
Gemini-1.5-flash    &  N/A  &	0.31       &	0.68        &	 0.45       &   0.35     &  0.27     &  0.34          &  2.68  \\

\midrule
\textbf{Open-Source }             &  &  &  &  &  & & &\\
Llama3.2          & 10.60B &	0.27       &	 N/A       &	0.39        &   0.18     &  0.11     &  0.24          &  4.82  \\

InternVL2.5-8B    & 8.08B  &	0.28       &	\textbf{0.63}        &	\textbf{0.50}        &   \textbf{0.26}     &   0.22    &   \textbf{0.31}         &  1.40  \\
InternVL2.5-4B    & 3.71B  &	\textbf{0.33}       &	0.57        &	0.42        &   0.20     &   \textbf{0.25}    &  0.30          & 0.88   \\
InternVL2.5-2B    & 2.21B  &	0.02       &	0.34        &	0.29        &   0.15     &   0.02    &  0.12          & 0.65   \\
InternVL2.5-1B    &  0.9B     &	0.09       &	0.25        &	0.17        &   0.01     &   0.14    &  0.10          & 0.51   \\

\midrule

Random              & N/A & 0.43 & 0.25 & 0.20 & 0.01 & 0.25 & 0.23 & N/A \\

\bottomrule
\end{tabular}
}
\end{center}
\end{table}

\begin{figure}[t!]
  \centering
  \includegraphics[width=\columnwidth]{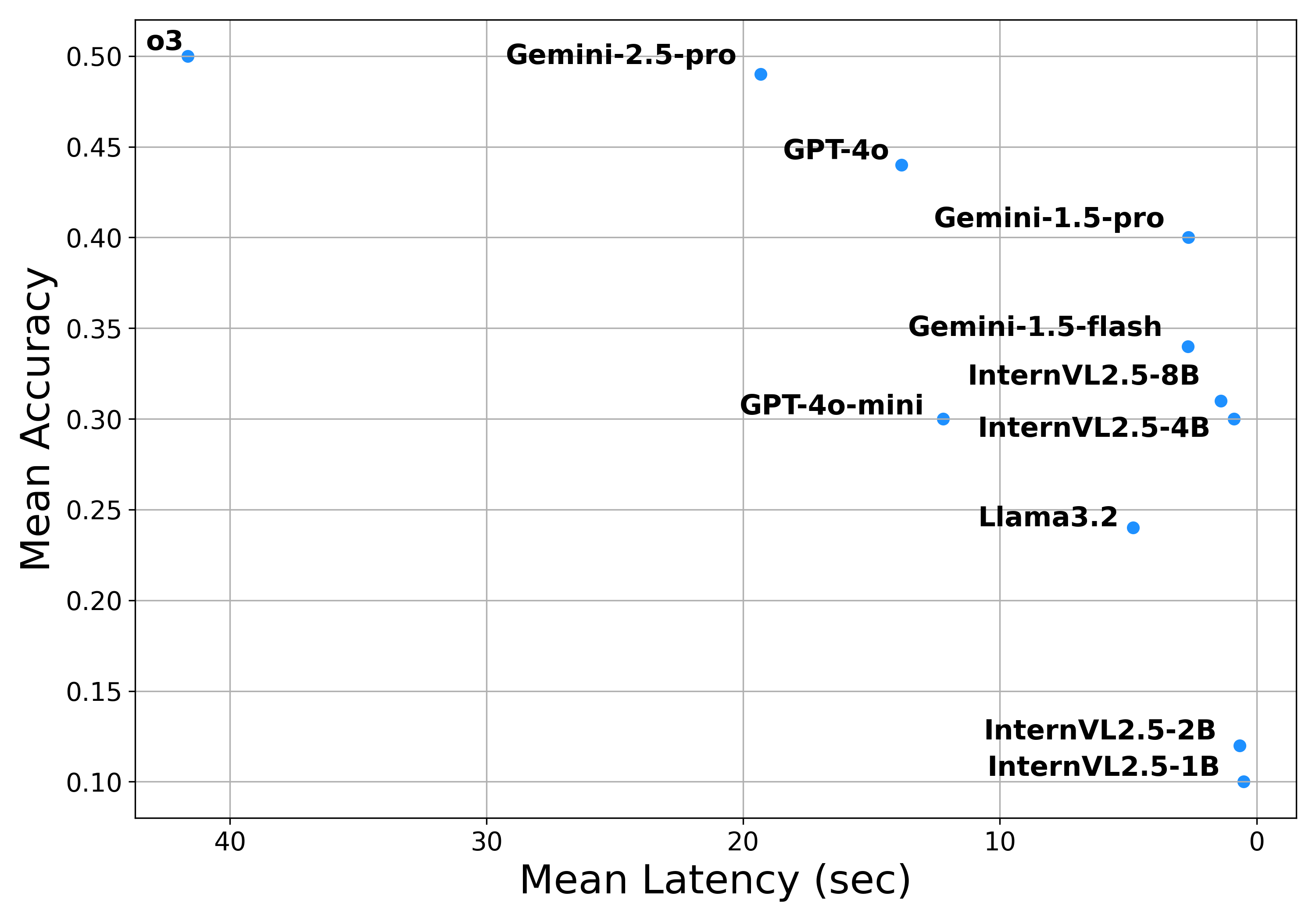}
    \caption{Performance-latency trade-off visualization across closed-source and open-source vision-language models (VLMs). We report average accuracy and latency (in seconds per question) across four HRIBench domains, excluding person identification due to model constraints, as discussed in Table~\ref{tab:results}.}
  \label{fig:tradeoff}
\end{figure}

Overall, as shown in Table~\ref{tab:results}, the highest performance on \benchmark was achieved by the closed-source \texttt{o3} and \texttt{Gemini-2.5-pro} models. This is consistent with findings from prior benchmarks showing that closed-source models were generally better at physical reasoning~\cite{chow2025physbench} and that \texttt{Gemini-2.5-pro} has strong robotics reasoning capabilities~\cite{zhao2025ManipBench}. 
Both \texttt{o3} and \texttt{Gemini-2.5-pro} leverage test-time computation, enabling them to significantly outperform other closed- and open-source models that lack this capability. However, this comes at the cost of high latency, making them unsuitable for real-time perception tasks. Despite their strong performance, a substantial gap remains between these models and the human baseline, indicating that they still fall short of the perceptual skills required for effective human-robot interaction. Among closed-source models without test-time reasoning, we observed some zero-shot generalization across HRI domains, but they also struggled with core perceptual challenges and latency. As a result, as shown in Fig.~\ref{fig:tradeoff}, although \texttt{Gemini-1.5-pro} achieved the best performance-latency tradeoff, none of the existing state-of-the-art closed-source models in our study demonstrated a satisfactory performance-latency trade-off suitable for real-time deployment. Prior research has shown that delays longer than 0.7 seconds disrupt the flow of natural human interaction~\cite{kendrick2015timing}.

To illustrate specific examples, as shown in Fig.~\ref{fig:example}, the green check mark denotes the ground truth, while the green \texttt{GPT-4o} logo indicates a mistake made by \texttt{GPT-4o}, and the Gemini logo indicates a mistake made by \texttt{Gemini-2.5-pro}.

In the domain of non-verbal cue understanding, models need to correlate hand gestures with a human’s eye gaze direction. Our results indicated that, while the models are already proficient at recognizing gestures alone, they struggled to understand fine-grained behaviors such as eye gaze direction, which is a crucial non-verbal cue for successful HRI.

In the domain of verbal instruction understanding, VLMs still struggled to reason about ambiguity and factual errors in verbal instructions. In the example shown in~Fig.~\ref{fig:example}, while a human evaluator can recognize the factual error (that the jacket the human needs is not on the table), the model may fail to reason about this and instead rigidly follows the instruction, selecting the most likely object on the table.

In the domains of human-robot-object relationship understanding and social navigation, our results indicated that VLMs also struggled with spatial reasoning skills that are crucial for HRI. For example, in the human-robot-object relationship domain (see Fig.~\ref{fig:example}), VLMs appeared to be incapable of reasoning from the human’s field of vision and often failed to distinguish between left and right directions. 

Similarly, in the domain of social navigation, analysis of the reasoning logs from \texttt{o3} and \texttt{Gemini-2.5-pro} suggested that these VLMs could successfully recognize different colored navigation paths, identify humans in the environment, and demonstrate a general understanding of what constitutes good social navigation. However, they frequently failed to comprehend the relative spatial relationships between the arrows and the humans.  As a result, they were often unable to select the correct answer. This limitation may stem from a fundamental lack of physical world understanding also observed in other recent VLM benchmark studies~\cite{chow2025physbench}.

Open-source VLMs can be run locally and reduce their parameter count to mitigate latency issues. However, our results empirically demonstrate that smaller model sizes resulted in a lack of fundamental perceptual capabilities needed for reliable deployment in real-time HRI systems. Among the evaluated models, the smaller~\texttt{InternVL2.5-8B} significantly out-performed the larger~\texttt{Llama3.2}. This result further highlights both the potential and the importance of future research into improving small-scale open-source VLMs to achieve better performance-latency trade-offs for HRI.

\section{Conclusion and Limitations}

In this work, we introduced \benchmark, a visual question-answering (VQA) benchmark with 1,000 novel questions designed to systematically evaluate vision-language models (VLMs) across a diverse set of human perceptual tasks critical for real-time human-robot interaction (HRI). \benchmark spans five key domains: (1) non-verbal cue understanding, (2) verbal instruction understanding, (3) human-robot-object relationship understanding, (4) social navigation, and (5) person identification. Our experiments show that none of the state-of-the-art (SOTA) closed- or open-source VLMs we evaluated demonstrate performance or latency that is suitable for real-time HRI. Furthermore, these models still struggle with fundamental human perceptual capabilities, particularly in understanding fine-grained multimodal cues such as eye gaze, ambiguous language-visual instructions, and real-world spatial and physical reasoning. These results highlight the need for future research to improve understanding of fine-grained behavioral cues and to develop smaller, low-latency VLMs capable of supporting real-time human perception in HRI.

\textbf{Limitations.} We acknowledge that while VLMs were evaluated independently for each question, human experts assessing \benchmark could have benefited from contextual information across questions, potentially improving the human baseline performance. However, due to input memory limitations, it is currently infeasible to include context from previous questions in the VLMs’ input. Additionally, the inference latency of closed-source models can vary based on external factors such as time of day and Internet connectivity, which may affect reproducibility. For open-source VLMs, all experiments were conducted on a machine equipped with two RTX 3090 GPUs, but latency may differ across hardware configurations. Finally, we conducted each experiment only once. Future work is needed to further evaluate the robustness and variability of model performance and latency across multiple runs.

\section{Acknowledgment}
This work was partially supported by the National Science Foundation (IIS-1925083), the National Institutes of Health (NIH-R01MH139134), the NSF CISE Graduate Fellowship (CSGrad4US) under Grant No. 2313998 (Award ID G-2A-061), and the Center for Undergraduate Research in Viterbi Engineering (CURVE) Fellowship at the University of Southern California. We also acknowledge Amy O’Connell for her contributions to curating data in the person identification domain. We used LLMs to help with proofreading our writing and manually verified all such content.



%
%
\bibliographystyle{spmpsci}
\bibliography{ref} 









\end{document}